\documentclass[letterpaper]{article} 

\ifdefined\aaaianonymous
    \usepackage[submission]{aaai2026}  
\else
    \usepackage{aaai2026}              
\fi

\usepackage{times}  
\usepackage{helvet}  
\usepackage{courier}  
\usepackage[hyphens]{url}  
\usepackage{graphicx} 
\usepackage{natbib}  
\usepackage{caption} 
\usepackage{algorithm}
\usepackage{algorithmic}

\usepackage{newfloat}
\usepackage{listings}
\DeclareCaptionStyle{ruled}{labelfont=normalfont,labelsep=colon,strut=off} 
\floatstyle{ruled}
\newfloat{listing}{tb}{lst}{}
\floatname{listing}{Listing}

\usepackage[dvipsnames]{xcolor}
\usepackage{pifont}
\usepackage{array}
\usepackage{booktabs}
\usepackage{multirow}

\nocopyright 

\ifdefined\aaaianonymous
    \title{TIME: Temporal-Sensitive Multi-Dimensional Instruction Tuning and Robust Benchmarking for Video-LLMs}
\else
    \title{TIME: Temporal-Sensitive Multi-Dimensional Instruction Tuning and Robust Benchmarking for Video-LLMs}
\fi

\author{
    Yunxiao Wang$^1$ \qquad Meng Liu$^2$ \qquad Wenqi Liu$^1$ \qquad Xuemeng Song$^3$ \qquad Bin Wen$^4$\\
    Fan Yang$^4$ \qquad Tingting Gao$^4$ \qquad Di Zhang$^4$ \qquad Guorui Zhou$^4$ \qquad Liqiang Nie$^1$\\
    \small{$^1$Shandong University} \qquad
    \small{$^2$Shandong Jianzhu University} \qquad
    \small{$^3$City University of Hong Kong} \qquad
    \small{$^4$Kuaishou Technology}\\
    {\tt\small \{yunxiao.wang, liuwq\}@mail.sdu.edu.cn}
    {\tt\small \{mengliu.sdu, sxmustc, nieliqiang\}@gmail.com}\\
    {\tt\small \{wenbin, yangfan, lisize, zhangdi08, zhouguorui\}@kuaishou.com}
}

\begin{document}

\maketitle

\begin{abstract}
Video large language models have achieved remarkable performance in tasks such as video question answering, however, their temporal understanding remains suboptimal. To address this limitation, we curate a dedicated instruction fine-tuning dataset that focuses on enhancing temporal comprehension across five key dimensions.  In order to reduce reliance on costly temporal annotations, we introduce a multi-task prompt fine-tuning approach that seamlessly integrates temporal-sensitive tasks into existing instruction datasets without requiring additional annotations. Furthermore, we develop a novel benchmark for temporal-sensitive video understanding that not only fills the gaps in dimension coverage left by existing benchmarks but also rigorously filters out potential shortcuts, ensuring a more accurate evaluation. Extensive experimental results demonstrate that our approach significantly enhances the temporal understanding of video-LLMs while avoiding reliance on shortcuts.

\end{abstract}

\ifdefined\aaaianonymous
\else


\section{Introduction}
\label{sec:intro}

Recent advancements in video large language models (video-LLMs)~\cite{tang2025video, cheng2025vilamp} have demonstrated significant capabilities in video understanding and reasoning. In contrast to image large language models (image-LLMs)~\cite{Kwai2025Keye-VL, coreteam2025mimovltechnicalreport}, which focus on static visual content analysis, video-LLMs must capture dynamic visual information across frames. This demands robust temporal understanding, a critical research challenge that has garnered substantial attention in recent work~\cite{Cheng2025V-STaR,Song2025Video-MMLU}.

\begin{figure}[htbp]
    \centering
    \includegraphics[width=1.0\linewidth]{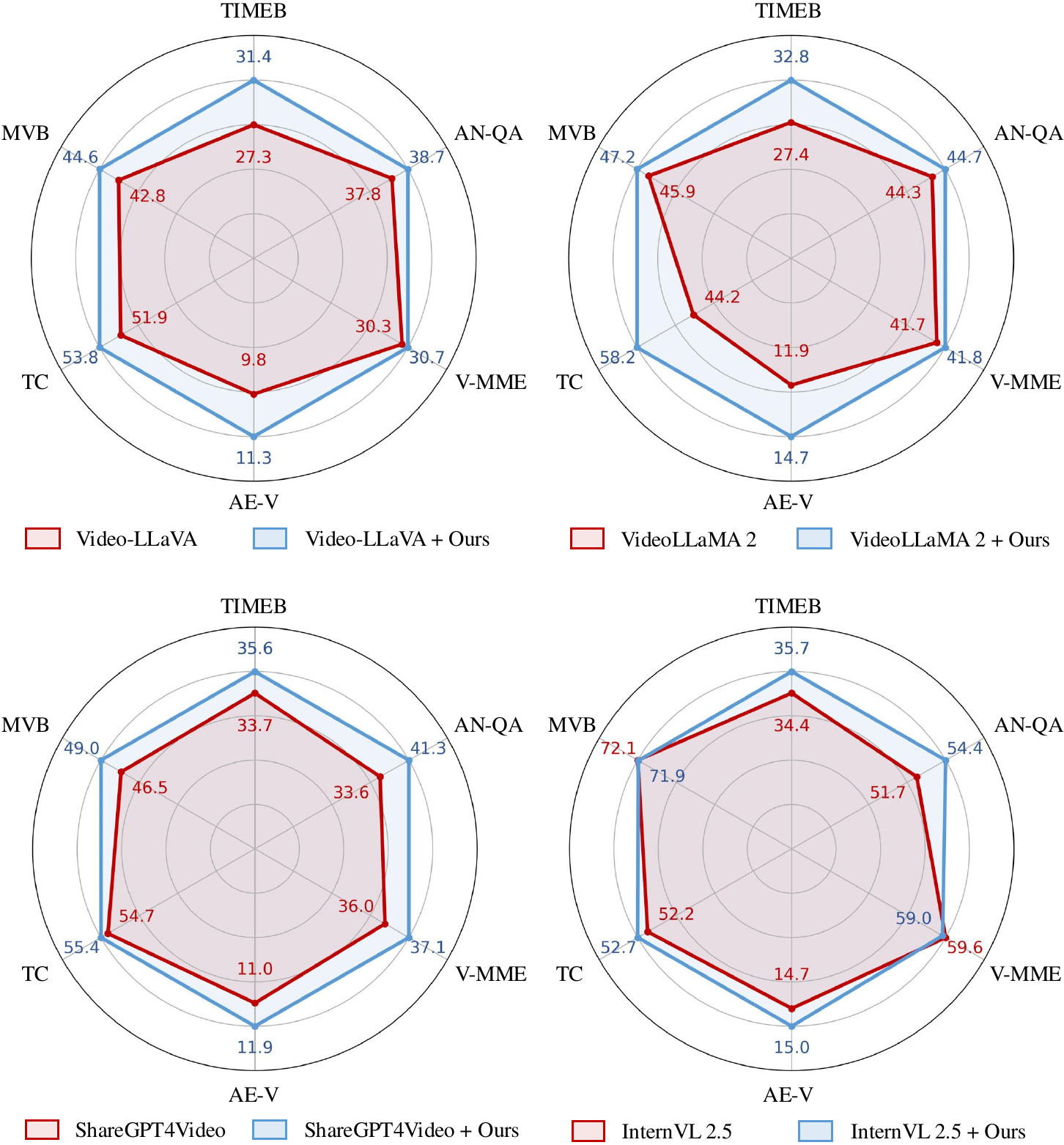}
    
    \caption{Performance of four video-LLMs evaluated across six benchmarks. ``+ Ours'' suffix denotes models fine-tuned with our approach, achieving substantial improvements on all benchmarks. TIMEBench (TIMEB), MVBench (MVB), TempCompass (TC), and AutoEval-Video (AE-V) focus on temporal scenarios, whereas Video-MME (V-MME) and ActivityNet-QA (AN-QA) evaluate general performance.}
    \label{fig:radar}
\end{figure}

Despite efforts to enhance temporal understanding in video-LLMs~\cite{Cheng2024VideoLLaMA, Liu2024ST-LLM}, recent studies indicate that significant challenges remain in tasks requiring advanced temporal reasoning~\cite{Wang2024VideoHallucer, Xiao2025VideoQA}. These limitations can be attributed to the following primary factors: \textbf{1) Insufficient temporal instruction fine-tuning data.} Current video instruction tuning datasets~\cite{Lin2024Video-LLaVA, Chen2024ShareGPT4Video} prioritize generalization across common scenarios rather than in-depth temporal comprehension. Although some approaches, such as AVicuna~\cite{Tang2024AVicuna}, generate temporal instruction tuning datasets through video splicing, their applicability remains limited to temporal localization due to the inherent constraints of this method. \textbf{2) Exploitation of data shortcuts.} For instance, in determining a child's movement direction (e.g., the \textit{Dynamic} dimension in Figure~\ref{fig:tasks}), models may rely on facial orientation instead of conducting a genuine temporal analysis of positional changes, leading to errors when videos are reversed. Notably, temporal benchmarks~\cite{Li2024MVBench,Liu2024TempCompass} are also affected by such shortcuts. Our experimental results (Figure~\ref{fig:benchmark}) show that performance based on single random frames significantly surpasses random baselines, indicating that many questions can be resolved without true temporal reasoning, thereby overestimating model capabilities.

\begin{figure*}[htbp]
    \centering
    \includegraphics[width=0.85\linewidth]{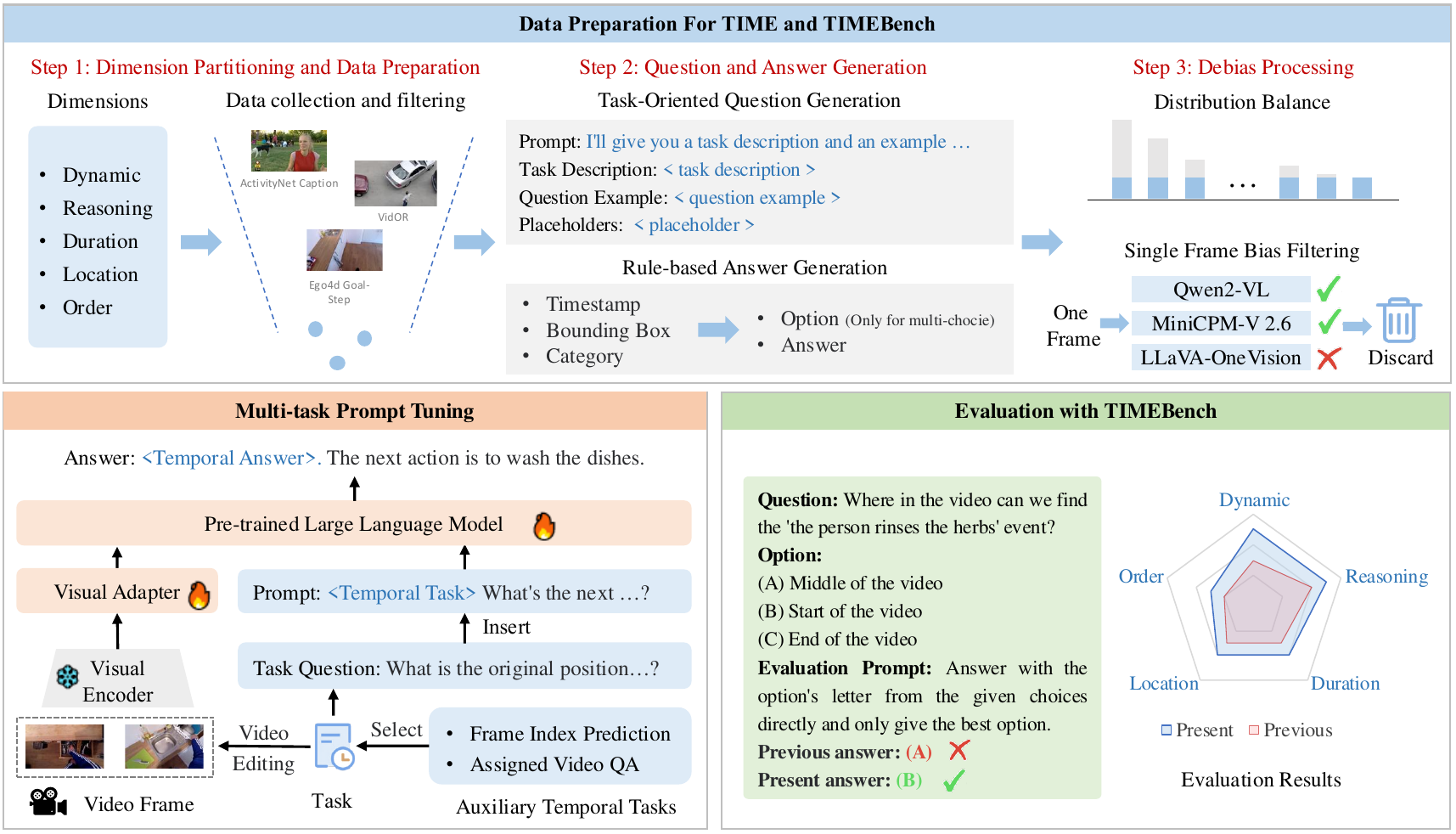}
    \caption{Overall framework overview. Our approach comprises three key components: the TIME instruction tuning dataset, the multi-task prompt tuning approach, and the TIMEBench video temporal understanding benchmark. The top panel shows the data preparation flow for TIME and TIMEBench in three sub-steps.  The bottom left illustrates the multi-task prompt tuning, where auxiliary temporal tasks are randomly interleaved with the original prompt, enabling the model to address additional temporal tasks alongside the main question.  The bottom right displays the evaluation format in TIMEBench, with all questions in a multiple-choice format for objective evaluation. }
    \label{fig:pipeline}
\end{figure*}

To address these challenges, we systematically enhance Video-LLMs' temporal understanding and evaluation through the following improvements. First, \textbf{we construct a Temporal-sensItive Multi-dimEnsional (TIME) instruction-tuning dataset} comprising 34,000 carefully curated samples across five key temporal dimensions (including \textit{Dynamic}, \textit{Reasoning}, \textit{Duration}, \textit{Location}, and \textit{Order}), which serves as a valuable complement to existing general-purpose instruction-tuning datasets. These dimensions capture essential aspects of temporal reasoning, such as the dynamics of events over time and the capacity to predict future occurrences. Simultaneously, we implement rigorous data filtering to remove any potential data shortcuts that might undermine the effectiveness of model tuning. Second, to further mitigate limitations in data volume, \textbf{we propose a Multi-Task Prompt tuning (MTP) framework} that augments standard instruction tuning with two types of temporal-oriented self-supervised tasks: frame-level task to improve single-frame temporal understanding, and segment-level task to enhance cross-segment temporal understanding. As shown in Figure~\ref{fig:radar}, fine-tuning with our approach results in substantial performance gains across most temporal benchmarks for the evaluated video-LLMs. Finally, \textbf{we develop a robust Temporal-sensItive Multi-dimEnsional Benchmark (TIMEBench)}, which covers five dimensions of temporal evaluation and provides a more accurate measure of temporal understanding ability by filtering out potential data shortcuts through a carefully designed filtering mechanism. Experimental results validate the rigor of our evaluation benchmark.

In summation, our contributions are as follows:
\begin{itemize}
    \item We construct a temporal-sensitive  multi-dimensional instruction-tuning dataset intended to augment existing general instruction-tuning corpora, incorporating curated samples that specifically address temporal reasoning challenges in video data.
    
    \item We introduce a novel multi-task prompt-tuning approach that enhances existing instruction fine-tuning processes with two temporal-oriented self-supervised tasks. 

    \item We propose a robust temporal-sensitive  multi-dimensional benchmark that provides a more accurate measure of temporal understanding ability by filtering out potential data shortcuts.

\end{itemize}

\begin{figure*}[htbp]
    \centering
    \includegraphics[width=0.9\linewidth]{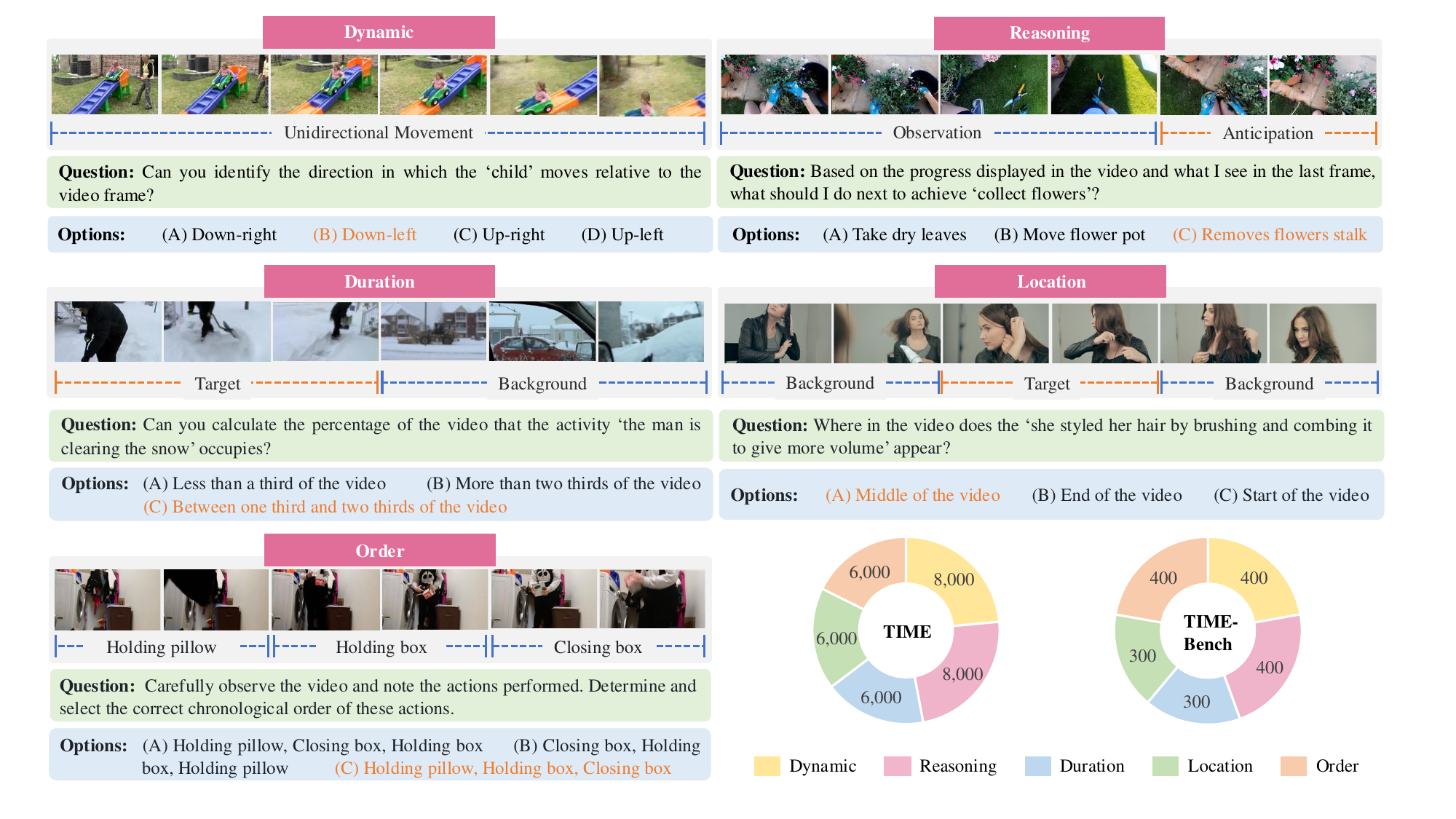}
    \caption{Examples from the TIME dataset and the TIMEBench benchmark with data distribution across task types. These examples encompass five tasks that target the \textit{Dynamic}, \textit{Reasoning}, \textit{Duration}, \textit{Location}, and \textit{Order} dimensions of temporal understanding. Ground-truth answers are highlighted in orange. In addition to multiple-choice questions, open-ended questions are also provided.}
    \label{fig:tasks}
\end{figure*}

\section{Temporal Sensitive Instruction Tuning}
\label{sec:method}
In this section, we present our approach to enhancing temporal understanding in video-LLMs through two key components: the construction of a temporally-sensitive instruction-tuning dataset and the development of a multi-task prompt fine-tuning strategy. These components are designed to augment the model's capacity for temporal reasoning across a wide range of video-related tasks.

\subsection{TIME Instruction Tuning Dataset}
The limited performance of existing video-LLMs on temporally sensitive tasks can be largely attributed to the absence of a dedicated instruction-tuning dataset tailored for temporal contexts. To address this gap, we construct the TIME instruction-tuning dataset, aimed at enhancing the model's temporal comprehension capabilities.

\subsubsection{Dimension and Data Selection}
To improve temporal comprehension in video-LLMs, we draw insights from related fields such as video moment localization~\cite{Liu2023Survey} and action anticipation~\cite{Hu2022Online}. We identify tasks that emphasize temporal understanding and categorize temporal reasoning into five key dimensions, as shown in Figure~\ref{fig:tasks}. Each dimension is paired with a task designed to strengthen the model's temporal reasoning capabilities. 

\noindent\textbf{Dynamic}. This dimension evaluates the model's ability to capture dynamic changes over time, specifically focusing on detecting the direction of object movement. Each video is cropped to highlight the unidirectional movement of a target object. Data is collected from the VidOR~\cite{Shang2019Annotating} dataset, originally designed for scene graph generation. To ensure unique identification, video segments containing more than two objects of the same category in a single frame are filtered out. The direction of the object's movement is determined by calculating the position of the center point of its bounding box. 

\noindent\textbf{Reasoning}. This dimension evaluates the model's capacity to forecast future events from observed temporal sequences. Specifically, we focus on action anticipation, where the model predicts the subsequent action based on preceding actions. For this task, we leverage the Ego4D Goal-Step~\cite{Song2023Ego4D} dataset, which provides annotations for goal-oriented hierarchical activities. In our framework, each step in the sequence is treated as an anticipatory action, while the earlier steps form the observation sequence. To ensure high-quality data, we filter out sequences with fewer than three steps and truncate those exceeding 15 steps. Additionally, only sequences with at least 50\% of the steps marked as ``essential'' are retained to mitigate noise.

\begin{figure*}[htbp]
    \centering
    \includegraphics[width=0.9\linewidth]{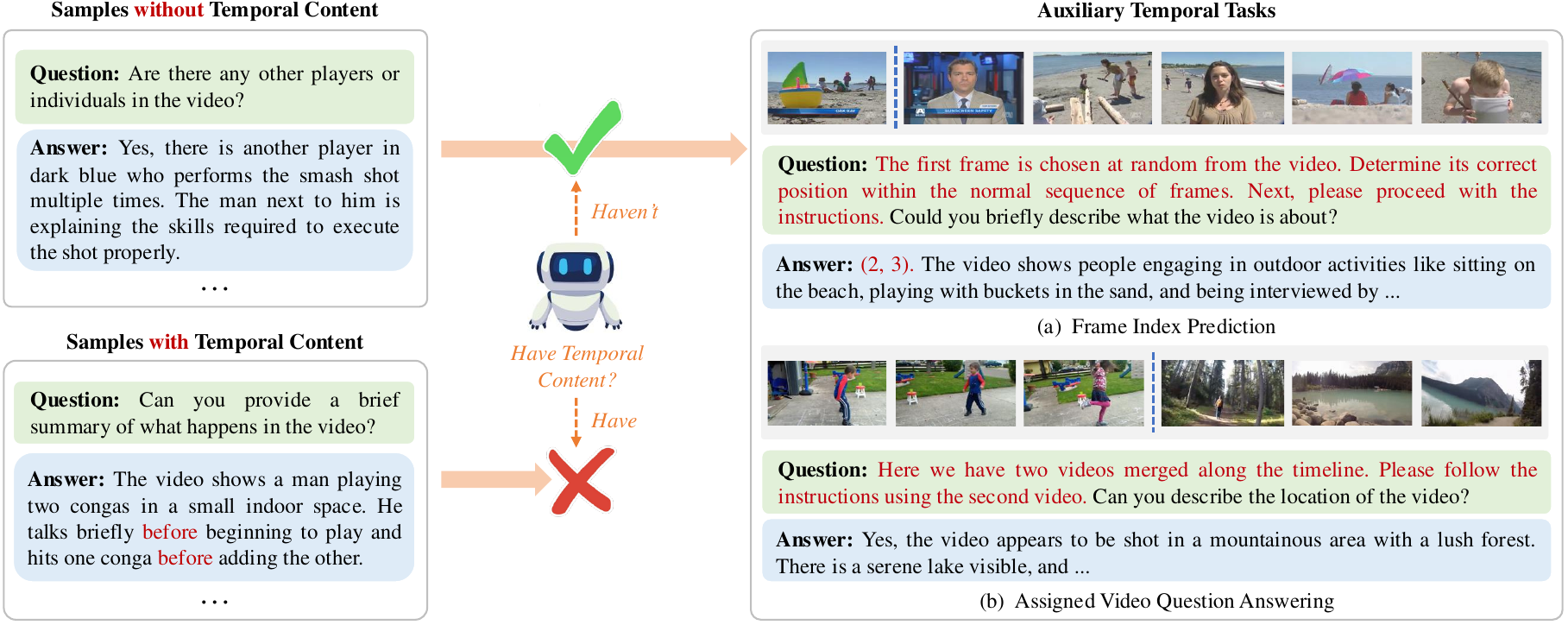}
    \caption{Illustration of our multi-task prompt tuning approach. The LLM first determines whether a sample contains a temporal description. Only samples lacking such descriptions are augmented with auxiliary tasks. In the frame index prediction task, a frame is randomly sampled from the sequence, and the model must restore its original position. In the assigned videoQA task, a randomly selected video is concatenated with the original, and the model needs to ignore the additional content to answer the original question.}
    \label{fig:multi}
\end{figure*}
\noindent\textbf{Duration}. This dimension evaluates the model's ability to perceive and estimate the duration of events in unedited videos. We challenge the model to determine the length of specific events by categorizing them into one of three segments based on their relative duration within the video. For this task, we use a subset of ActivityNet Captions~\cite{Krishna2017Dense} where events are divided into three parts according to their temporal span, and the model must correctly identify the appropriate category for each event.

\noindent\textbf{Location}. This dimension measures the model's capability to pinpoint precise temporal segments within a video by determining when a given event occurs, distinct from assessing its duration.
To construct the dataset for this task, we extract a non-overlapping subset from ActivityNet Captions~\cite{Krishna2017Dense} and uniformly partition each video into three intervals: start, middle, and end. Only activities that both begin and end within the same interval are selected, ensuring the model concentrates on accurately identifying temporal boundaries.

\noindent\textbf{Order}. This dimension assesses the model's ability to understand temporal event sequences by ordering actions accurately. Specifically, the model must arrange multiple non-overlapping actions in the sequence in which they occur. For this task, we use the Charades~\cite{Sigurdsson2016Hollywood} dataset, partitioning each video into segments comprising three distinct actions performed sequentially. To ensure sequence diversity, segments with identical actions are filtered out so that each sequence consists of unique action types. To ensure sufficient adequate context, we filter out video clips with very short target activities. This step reduces the likelihood of the model erring due to insufficient observational context rather than a lack of temporal understanding.

\subsubsection{Question and Answer Generation}
Since the original datasets lack pre-existing QA pairs, we generate them to build a comprehensive QA dataset. To maximize diversity, we partition the dataset into two parts: open-ended and multiple-choice QA.

\noindent\textbf{Open-ended QA}. 1) \textit{Question Generation}: For each task, we manually create an example question and then use ChatGPT~\cite{OpenAI2024GPT-4} to generate 10 variations based on the task description and example. In questions derived from the VidOR~\cite{Shang2019Annotating}, ActivityNet Captions~\cite{Krishna2017Dense}, and Ego4D Goal-Step~\cite{Song2023Ego4D} datasets, placeholders are inserted and later replaced with specific target objects, activities, or goal descriptions. 2) \textit{Answer Generation}: For \textit{Dynamic} and \textit{Reasoning}, the correct answer typically reflects the direction of movement or the action category. For \textit{Duration} and \textit{Location}, answers are derived from event-associated timestamps,  while for \textit{Order}, the answer is the chronological arrangement of three actions.

\noindent\textbf{Multiple-choice QA}. 1) \textit{Question Generation:} We follow the similar process as for open-ended QA, but incorporate additional instructions and answer options. 2) \textit{Instruction Generation}: For each task, we generate 10 extra instructional prompts using ChatGPT~\cite{OpenAI2024GPT-4}. These prompts, inserted before the question, guide video-LLMs to select the correct option from a list of candidates.
 3) \textit{Option Generation}: 
For \textit{Dynamic}, \textit{Duration}, and \textit{Location}, all possible answers are provided as options. For \textit{Order}, random permutations of actions are used to avoid shortcuts based on visual or keyword co-occurrence.
For \textit{Reasoning}, incorrect options are sampled from other steps within the same goal to minimize dependence on co-occurrence relationships.

\subsubsection{Dataset Debiasing}

Prior studies~\cite{Tong2024Eyes,Yu2024HalluciDoctor,Zhang2022Incorporating,Huang2024OPERA} have shown that multimodal language models are prone to biases arising from heuristics such as keyword co-occurrence and inherent distributional patterns in training data, which can lead to hallucinatory behavior in video-LLMs. For example, Otani et al.~\cite{Otani2020Uncovering} observed that many video moment location datasets exhibit a temporal bias, with events predominantly occurring at the beginning of the video. Consequently, models may over-predict early events, achieving high accuracy on biased test sets without genuine temporal comprehension.

To mitigate these biases and foster a true understanding of temporal events, we implement the following debiasing strategies: 1) We initially conduct a manual verification of the content generated by ChatGPT~\cite{OpenAI2024GPT-4} to ensure its accuracy. Then, we ensure a balanced distribution of answers across both open-ended and multiple-choice QA formats. Specifically, for \textit{Dynamic}, \textit{Duration}, and \textit{Location}, we maintain roughly equal counts of each answer type to prevent skewed representations. 2) For \textit{Reasoning} and \textit{Order}, we downsample frequently occurring actions to alleviate long-tailed distributions. 3) For \textit{Dynamic}, we incorporate a reversed version of the video to discourage reliance on visual shortcuts, such as face or vehicle orientation. And 4) we carefully balance the distribution of questions and options across all temporal dimensions.

\subsection{Multi-task Prompt Tuning}
Relying solely on fully annotated datasets for instruction tuning is inherently limited by annotation availability. While existing work~\cite{Lei2021Understanding,Wang2023All,Sun2022Long} has introduced unsupervised tasks such as masked frame modeling to improve temporal understanding, these approaches are not directly applicable to fine-tuning video-LLMs. To overcome this limitation, we propose a multi-task prompt-tuning approach that incorporates two auxiliary tasks into existing instruction-tuning datasets without requiring additional annotations.

\noindent\textbf{Frame Index Prediction.} In this task, we randomly sample a frame from the original video and position it at the beginning of the sequence. A prompt is inserted before the original question to guide the model in predicting the correct position of the displaced frame.  The modified question-answer structure is illustrated in the upper-right of Figure~\ref{fig:multi}.

\noindent\textbf{Assigned VideoQA.} This task trains the model to identify the relevant video segments required to answer a given question, similar to the location task. Here, a randomly selected video from the dataset is concatenated with the original video, with the order of the videos randomized. Because the initial questions focus on video content descriptions without detailed location cues, explicit prompt instructions are provided to help the model discern the correct video, without dictating the proportion of the two videos. This setup is shown in the bottom-right of Figure~\ref{fig:multi}.

In both tasks, the temporal content of the original video is minimally altered to ensure that the model can still accurately answer the original question.  Moreover, as depicted in Figure~\ref{fig:multi}, we employ Qwen2.5-72B~\cite{qwen2.5}, an open-source LLM with capabilities comparable to ChatGPT~\cite{OpenAI2024GPT-4}, to identify QA pairs that involve temporal content. We refrain from adding new tasks to these pairs to avoid introducing errors from video modification. For each auxiliary task, 10 prompts are generated using ChatGPT~\cite{OpenAI2024GPT-4}. Following the workflow in Figure ~\ref{fig:pipeline}, we tune the video-LLMs with these auxiliary tasks. Specifically, for samples without temporal content, one auxiliary task is randomly selected, the video and QA pair are modified accordingly, and then the video-LLMs are tuned on the rest of the dataset. This approach opens a new avenue for unsupervised instruction fine-tuning aimed at enhancing the temporal comprehension abilities of LLMs.

\section{TIMEBench}
\label{sec:bench}
In this section, we introduce TIMEBench, a novel benchmark designed to evaluate temporal understanding in video models, and compare it with existing benchmarks. TIMEBench comprehensively assesses the temporal reasoning capabilities of video models across five dimensions: \textit{Dynamic}, \textit{Reasoning}, \textit{Duration}, \textit{Location}, and \textit{Order}.  Following the pipeline illustrated in  Figure~\ref{fig:pipeline}, we systematically collect and filter data from diverse sources and generate multiple-choice questions for objective evaluation.  This design not only mitigates data shortcuts but also extends the evaluation to a broader range of temporal aspects, setting TIMEBench apart from existing benchmarks.

\subsection{Benchmark Construction}
As discussed previously, existing benchmarks often fall short of covering the five dimensions outlined in 
Section~\ref{sec:method} and are susceptible to shortcut issues. To overcome these shortcomings, we develop a benchmark that follows the data preparation and processing pipeline described in Section~\ref{sec:method}.
The key differences are outlined below.

\begin{figure}[t]
    \centering
    \includegraphics[width=\linewidth]{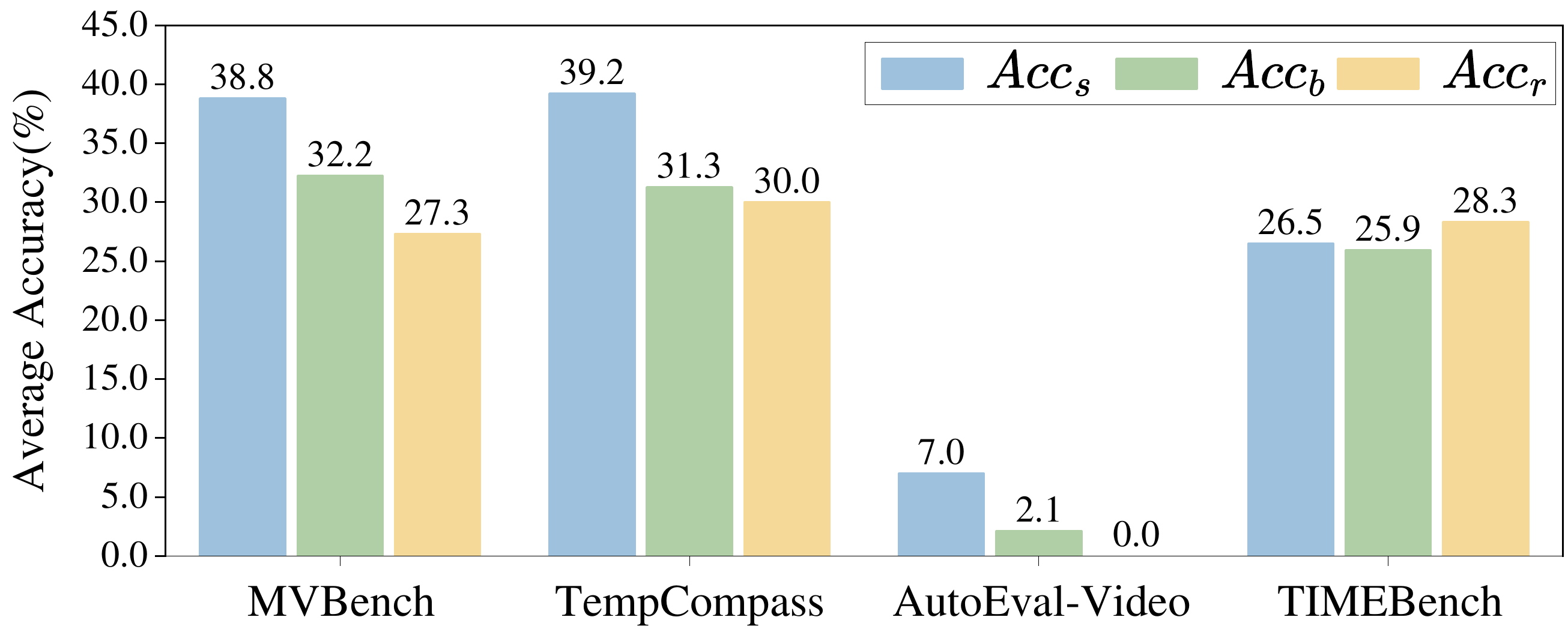}
    \caption{Performance comparison of VideoLLaMA~2~\cite{Cheng2024VideoLLaMA} on existing Benchmarks. $Acc_s$ represents the predictions made using only a single frame, $Acc_b$ denotes the accuracy when visual information is excluded, and $Acc_r$ indicates random predictions.}
    \label{fig:benchmark}
\end{figure}

\begin{table*}[t]
\centering


\begin{tabular}{ccccccccccccc}
\toprule
\multirow{2}{*}{Model} & \multicolumn{6}{c}{TIMEBench} & \multirow{2}{*}{MVB}  & \multirow{2}{*}{TC} & \multirow{2}{*}{AE-V} & \multirow{2}{*}{V-MME} & \multirow{2}{*}{AN-QA} \\ 
\cmidrule(lr){2-7}
 & LO & DU & DY & OR & RE & AVG & & \\
\midrule
Random   & 33.3 & 33.3 & 25.0 & 25.0 & 25.0 & 28.3& 27.3  & 30.0 & 0.0 & 27.2 & 0.0 \\
\midrule

Video-LLaVA      & 32.7 & 32.3 & 22.8 & 26.5 & 24.8 & 27.3 & 42.8 & 51.9 & 9.8 & 30.3 & 37.8 \\ 
+ Ours                                   & \textbf{37.0}  & \textbf{34.3}  & \textbf{25.5}  & \textbf{34.3}  & \textbf{28.3}  & \textbf{31.4}  & \textbf{44.6}    & 
\textbf{53.8} & \textbf{11.3} & \textbf{30.7} & \textbf{38.7} \\ 

\midrule
VideoLLaMA~2     & 32.3 & 30.3 & 22.8 & 28.0 & 24.0 & 27.4 & 45.9 & 44.2 & 11.9 & 41.7 & 44.3 \\
+ Ours                                 & \textbf{33.3} & \textbf{43.0} & \textbf{28.0} & \textbf{33.0} & \textbf{26.8} & \textbf{32.8} & \textbf{47.2} & \textbf{58.2} & \textbf{14.7} & \textbf{41.8} & \textbf{44.7} \\
\midrule
ShareGPT4Video &  33.7  &  46.7  & 31.0 &  33.8  &  26.5 & 33.7 & 46.5 & 54.7 & 11.0 & 36.0 & 33.6 \\ 
+ Ours   &  \textbf{38.3}  &  \textbf{47.3}  &  \textbf{31.7} &  \textbf{34.5} &  \textbf{29.5} &  \textbf{35.6} & \textbf{49.0} & \textbf{55.4} & \textbf{11.9} & \textbf{37.1} & \textbf{41.3} \\
\midrule
InternVL~2.5 & 30.3 & 45.7 & 29.3 & 42.0 & 26.8 & 34.4 & \textbf{72.1} & 52.2 & 14.7 & \textbf{59.6} & 51.7\\
+ Ours    & \textbf{30.6} & \textbf{48.3} & \textbf{29.8} & \textbf{42.8} & \textbf{27.0} & \textbf{35.7} & 71.9 & \textbf{52.7} & \textbf{15.0} & 59.0  & \textbf{54.4}\\
\bottomrule
\end{tabular}

\caption{Performance comparison across six benchmarks. The best performance for each part is highlighted in \textbf{bold}. TIMEBench, MVB, TC and AE-V focus on temporal scenarios, whereas V-MME and AN-QA address general scenario. TIMEBench, MVB and V-MME are multiple-choice forms, AE-V and AN-QA are open-ended forms, and TC contains both multiple-choice and open-ended forms.}
\label{tab:performance}
\end{table*}

\noindent\textbf{Data Preparation.} 
We leverage diverse data sources to construct TIMEBench, ensuring robust evaluation on out-of-domain data.
Although both Ego4D Goal-Step~\cite{Song2023Ego4D} and EgoPlan~\cite{Chen2024EgoPlan} originate from Ego4D~\cite{Grauman2022Ego4D}, we exclusively use the subset of EgoPlan sourced from Epic-Kitchens~\cite{Damen2022Rescaling} to avoid domain overlap. For the Breakfast~\cite{Kuehne2014The}, we use coarse-level annotations to avoid overly fine-grained action partitioning, which could impede the model’s ability to distinguish actions.
Furthermore, for TACoS~\cite{Regneri2013Grounding} and QVHighlights~\cite{Lei2021Detecting}, we apply random video cropping and generate question-answer options based on the new timestamps, thereby minimizing reliance on prior knowledge or fixed temporal context.

\noindent\textbf{QA generation.} Following the pipeline in Section~\ref{sec:method}, we construct QA pairs with an emphasis on multiple-choice formats to ensure objective evaluation. The evaluation prompt is defined as: ``Answer with the option's letter from the given choices directly and only give the best option''. A complete example is provided in the bottom right of Figure~\ref{fig:pipeline}. 

\noindent\textbf{Benchmark Debiasing.} Prior work~\cite{Tong2024Eyes} as demonstrated that current models often exploit data shortcuts rather than exhibiting true temporal comprehension. To address this, we utilize three advanced open-source multimodal LLMs, i.e., Qwen2-VL~\cite{Wang2024Qwen2-VL}, LLaVA-OneVision~\cite{Li2025LLaVA-OneVision}, and MiniCPM-V 2.6~\cite{Yao2024MiniCPM-V}, as judge agents to evaluate the validity of QA pairs. 
Specifically, each question is based on a randomly selected frame from the video, and the models are tasked with answering using only that single frame. Given the limited temporal comprehension capabilities of current video-LLMs~\cite{Wang2024VideoHallucer,Li2024MVBench}, we adopt a majority voting approach. If two or more judge models answer correctly based solely on a single sampled frame, this indicates potential shortcut reliance, and the corresponding QA pair is filtered out. The final benchmark is constructed from the remaining valid pairs, and resampling is performed to ensure a balanced distribution of answer options.

\subsection{Benchmark Comparison}

To ensure the objectivity of our evaluation, we carefully debias the construction of TIMEBench. To assess the effectiveness of this approach,
we conducted an experiment (see Figure~\ref{fig:benchmark}) comparing the impact of potential biases across various benchmarks. In this experiment, we use the random prediction accuracy $Acc_r$ as a baseline and report $Acc_s$ and $Acc_b$ for comparison. $Acc_s$ represents the accuracy of VideoLLaMA~2~\cite{Cheng2024VideoLLaMA} using a single video frame as input, while $Acc_b$ corresponds to the blind setting where videos are replaced with fully black images.

The results clearly demonstrate that on MVBench \cite{Li2024MVBench}, TempCompass \cite{Liu2024TempCompass}, and AutoEval-Video \cite{Chen2023AutoEval}, the $Acc_b$ consistently surpasses $Acc_r$ even in the absence of visual information. Moreover, with the addition of just one frame from the videos, $Acc_s$ on these three benchmarks significantly exceeds both $Acc_r$ and $Acc_b$, indicating that many questions in these benchmarks can be answered without relying on temporal information. In contrast, on TIMEBench, both $Acc_s$ and $Acc_b$ fall below $Acc_r$, with a small gap between them, indicating that TIMEBench is less prone to shortcut biases and offers a more rigorous evaluation of temporal understanding. It can be attributed to our meticulous data filtering and debiasing procedures.

\begin{table*}[t]
\centering


\begin{tabular}{ccccccccccccc}
\toprule
\multirow{2}{*}{Model} & \multicolumn{6}{c}{TIMEBench} & \multirow{2}{*}{MVB}  & \multirow{2}{*}{TC} & \multirow{2}{*}{AE-V} & \multirow{2}{*}{V-MME} & \multirow{2}{*}{AN-QA} \\
\cmidrule(lr){2-7}
 & LO & DU & DY & OR & RE & AVG & & \\
\midrule
Baseline  & 32.3 & 30.3 & 22.8 & 28.0 & 24.0 & 27.4 & 45.9 & 44.2 & 11.9 & 41.7 & 44.3 \\

+ TIME    & \underline{34.3} & \underline{41.0} & \underline{26.5} & \textbf{34.8} & \underline{26.3} & \underline{32.6} & \underline{47.1} & \underline{57.4} & \underline{14.0} & 41.6 & \underline{44.6} \\ 

+ MTP     & \textbf{35.3} & 32.0 & 24.5 & 31.0 & 24.8 & 29.5 & 45.9 & 49.7 & 13.8 & \textbf{41.9} & 44.5 \\

+ ALL     & 33.3 & \textbf{43.0} & \textbf{28.0} & \underline{33.0} & \textbf{26.8} & \textbf{32.8} & \textbf{47.2} & \textbf{58.2} & \textbf{14.7} & \underline{41.8} & \textbf{44.7} \\ 
\bottomrule
\end{tabular}

\caption{Ablation study of TIME dataset and MTP method. The best performance is highlighted in \textbf{bold}, and the second-best result is indicated with \underline{underlines}.}
\label{tab:ablation}
\end{table*}

\section{Experiments}

\noindent\textbf{Experiment Settings.} For evaluation, we used TIMEBench and five additional benchmarks: MVBench~\cite{Li2024MVBench}, TempCompass~\cite{Liu2024TempCompass} and AutoEval-Video~\cite{Chen2023AutoEval}, which are tailored for video temporal understanding, as well as Video-MME~\cite{Fu2025Video-MME} and ActivityNet-QA~\cite{Yu2019ActivityNet}, which serve as general benchmarks for video comprehension. To validate the effectiveness of our method, we fine-tuned Video-LLaVA~\cite{Lin2024Video-LLaVA}, VideoLLaMA~2~\cite{Cheng2024VideoLLaMA}, ShareGPT4Video~\cite{Chen2024ShareGPT4Video} and InternVL~2.5~\cite{chen2024internvl} using our TIME dataset and our MTP approach\footnote{For Video-LLaVA, VideoLLaMA~2 and ShareGPT4Video, we combined their respective original instruction datasets with our TIME dataset. In the case of InternVL~2.5, since the the original instruction dataset is unavailable, we directly fine-tuned the model using our TIME dataset on the pre-trained checkpoint.}. Specifically, for Video-LLaVA and VideoLLaMA~2, the entire parameters of the LLMs was fine-tuned, whereas for ShareGPT4Video and InternVL~2.5, we employed LoRA to fine-tuning. Except for ShareGPT4Video, the visual encoders were frozen for all models. Further details are in the supplementary material.

\noindent\textbf{Performance Comparison.}
As illustrated in Table~\ref{tab:performance}, fine-tuning with our method leads to significant improvements for all video-LLMs across most benchmarks, especially on the four benchmarks that specifically assess video temporal understanding. 
Simultaneously, performance on the two general benchmarks remains stable or shows slight improvements, indicating that our method enhances temporal reasoning without compromising overall performance. Notably, the results on TIMEBench are closer to random performance, underscoring the stricter evaluation criteria of our benchmark in assessing temporal dimensions.

\begin{table}[t]
\centering

\begin{tabular}{ccccccc}
\toprule
Method  & LO & DU & DY & OR & RE & AVG \\
\midrule
Baseline & 32.3 & 30.3 & 22.8 & 28.0 & 24.0 & 27.4 \\
Mixing  & \textbf{34.3} & \textbf{41.0} & \textbf{26.5} & \textbf{34.8} & \textbf{26.3} & \textbf{32.6} \\
After   & 31.3 & 30.3 & 23.0 & 25.3 & \underline{25.3} & 26.6 \\
Before  & \underline{34.2} & \underline{35.3} & \underline{23.3} & \underline{29.8} & 24.0 & \underline{28.7} \\
\bottomrule
\end{tabular}

\caption{Ablation study on the strategy for incorporating the TIME dataset. The best performance is highlighted in \textbf{bold}, and the second-best result is indicated with \underline{underlines}.}
\label{tab:time}
\end{table}
\begin{table}[t]
\centering

\begin{tabular}{ccccccc}
\toprule
Method  & LO & DU & DY & OR & RE & AVG  \\
\midrule
100\% & 32.7 & 30.3 & 25.5 & \underline{29.5} & 23.3 & 27.9 \\
75\%  & 34.0 & \underline{31.0} & \underline{26.3} & 28.0 & 22.5 & 27.9 \\
50\%  & \underline{35.3} & 30.3 & 22.5 & \textbf{30.5} & \textbf{27.5} & \underline{28.8} \\
25\%  & \textbf{38.0} & \textbf{31.3} & \textbf{26.5} & 26.5 & \underline{25.0} & \textbf{28.9} \\
0\% & 32.3 & 30.3 & 22.8 & 28.0 & 24.0 & 27.4 \\
\bottomrule
\end{tabular}

\caption{Ablation study on frame index prediction task ratio.}
\label{tab:frame}
\end{table}
\noindent\textbf{Ablation Study.}
We conducted ablation experiments on VideoLLaMA~2~\cite{Cheng2024VideoLLaMA} to dissect the contributions of our approach. Table~\ref{tab:ablation} shows that both TIME and MTP independently improve the model's temporal understanding, while their combination achieves the best balance between temporal tasks and general tasks. 

In Table~\ref{tab:time}, we compared different strategies for integrating TIME data into the training process.  Our findings indicate that mixing TIME data with the original instruction fine-tuning dataset concurrently yields superior performance compared to sequentially adding TIME data after or before the original fine-tuning. This improvement is likely due to mitigating the distribution gap between temporal tasks and general tasks by integrating them together.

Table~\ref{tab:frame} and Table~\ref{tab:two} explore the impact of different proportions of auxiliary temporal tasks. We observe that introducing any proportion of auxiliary tasks improves performance on temporal tasks, confirming their effectiveness. However, increasing the proportion of auxiliary tasks beyond a certain threshold does not yield further improvements and may even degrade performance. Based on our experiments,  the optimal mix is 50\% assigned video question-answering tasks and 25\% frame index prediction tasks.

\begin{table}[t]
\centering

\begin{tabular}{ccccccc}
\toprule
Method  & LO & DU & DY & OR & RE & AVG \\
\midrule
100\% & \underline{35.0} & \underline{35.3} & 25.3 & \underline{32.0} & \underline{25.4} & \underline{30.1} \\
75\%  & \textbf{36.7} & 31.0 & 26.3 & 30.8 & 24.3 & 29.3 \\
50\%  & 34.3 & \textbf{41.0} & \textbf{27.5} & \textbf{32.8} & \textbf{25.5} & \textbf{31.6} \\
25\%  & 33.7 & 30.7 & \underline{26.8} & 29.5 & 23.8 & 28.5 \\
0\%   & 32.3 & 30.3 & 22.8 & 28.0 & 24.0 & 27.4 \\
\bottomrule
\end{tabular}

\caption{Ablation study on the assigned videoQA task ratio.}
\label{tab:two}
\end{table}

\begin{figure}[t]
    \centering
    \includegraphics[width=\linewidth]{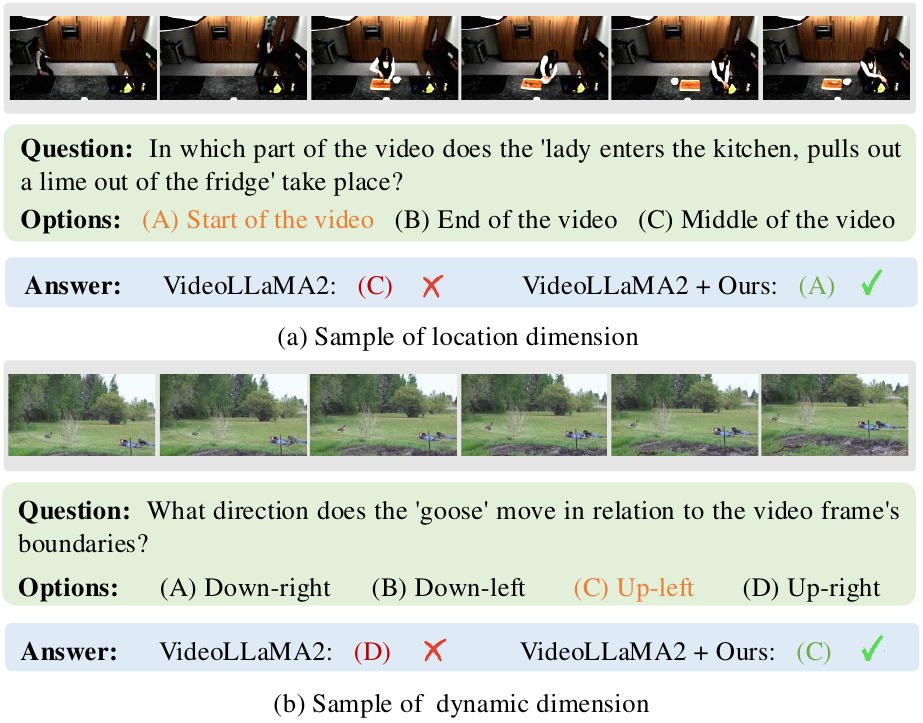}
    \caption{Visualization of partial sample prediction results in TIMEBench. The correct answer is highlighted in orange.}
    \label{fig:qualitative}
\end{figure}

\noindent\textbf{Qualitative Analysis.}
Figure~\ref{fig:qualitative} compares the predictions of the original VideoLLaMA~2~\cite{Cheng2024VideoLLaMA} and the fine-tuned version on two TIMEBench samples. In the \textit{Location} example, the fine-tuned model better captures subtle temporal location differences, while in the \textit{Dynamic} sample (with reversed video), the fine-tuned model relies less on priors like face orientation compared to the original model.

\section{Conclusion}
\label{sec:conclusion}
In this work, we present a comprehensive framework to enhance temporal understanding in video-LLMs by delineating five key dimensions of temporal reasoning. Building on these dimensions, we introduced the TIME instruction fine-tuning dataset and a novel multi-task prompt-tuning approach that overcomes the limitations of traditional data labeling. Furthermore, our proposed TIMEBench benchmark exposes the temporal shortcomings of existing video-LLMs while validating the effectiveness of our approach in advancing temporal reasoning capabilities.

\bibliography{aaai2026}
\end{document} 

\typeout{get arXiv to do 4 passes: Label(s) may have changed. Rerun}